\documentclass{article}
\pdfoutput=1

\usepackage{times}
\usepackage{graphicx} % more modern
\usepackage{subfigure} 

\usepackage{natbib}

\usepackage{algorithm}
\usepackage{algorithmic}

\usepackage{hyperref}

\usepackage[accepted]{icml2016} 

\icmltitlerunning{Semi-Supervised Learning with Generative Adversarial Networks}

\begin{document} 

\twocolumn[
\icmltitle{Semi-Supervised Learning with Generative Adversarial Networks}

\icmlauthor{Augustus Odena}{augustus.odena@gmail.com}

\icmlkeywords{ICML}

\vskip 0.3in
]

\begin{abstract} 
We extend Generative Adversarial Networks (GANs) to the semi-supervised context by forcing the discriminator network to output class labels.
We train a generative model G and a discriminator D on a dataset with inputs belonging to one of N classes.
At training time, D is made to predict which of N+1 classes the input belongs to, where an extra class is added to correspond to
the outputs of G.
We show that this method can be used to create a more data-efficient classifier
and that it allows for generating higher quality samples than a regular GAN.
\end{abstract} 

\section{Introduction}
\label{submission}

Work on generating images with Generative Adversarial Networks (GANs) has shown promising results \cite{GANS}.
A generative net G and a discriminator D are trained simultaneously with conflicting objectives.
G takes in a noise vector and outputs an image, while
D takes in an image and outputs a prediction about whether the image is a sample from G.
G is trained to maximize the probability that D makes a mistake, and D is trained to minimize that probability.
Building on these ideas, one can generate good output samples using a cascade \cite{LAPGAN} of convolutional neural networks.
More recently \cite{DCGAN}, even better samples were created from a single generator network.
Here, we consider the situation where we try to solve a semi-supervised classification task and learn a generative model simultaneously.
For instance, we may learn a generative model for MNIST images while we train an image classifier, which we'll call C.
Using generative models on semi-supervised learning tasks is not a new idea - Kingma et al. \yrcite{SSVAE} expand work on variational generative techniques 
\cite{VAES, VAES2} to do just that.
Here, we attempt to do something similar with GANs.
We are not the first to use GANs for semi-supervised learning.
The CatGAN \cite{CATGAN} modifies the objective function to
take into account mutual information between observed examples and their predicted
class distribution.
In Radford et al. \yrcite{DCGAN}, the features learned by D are reused in classifiers.

The latter demonstrates the utility of the learned representations, but it has several undesirable properties.
First, the fact that representations learned by D help improve C is not surprising - it seems
reasonable that this should work. However, it also seems reasonable that learning a good C would help to improve the performance of D.
For instance, maybe images where the output of C has high entropy are more likely to come from G.
If we simply use the learned representations of D
after the fact to augment C, we don't take advantage of this.
Second, using the learned representations of D after the fact doesn't allow for training C and G simultaneously.
We'd like to be able to do this for efficiency reasons, but there is a more important motivation.
If improving D improves C, and improving C improves D (which we know improves G) then we may be able to
take advantage of a sort of feedback loop, in which all 3 components (G,C and D) iteratively make each other
better.

In this paper, inspired by the above reasoning, we make the following contributions:

\begin{itemize}
  \item First, we describe a novel extension to GANs that allows them to learn a generative model and a classifier simultaneously.
We call this extension the Semi-Supervised GAN, or SGAN.
\item Second, we show that SGAN improves classification performance on restricted data sets over a baseline classifier with no generative component.
\item Finally, we demonstrate that SGAN can significantly improve the quality of the generated samples and reduce
  training times for the generator.
\end{itemize}

\section{The SGAN Model}

The discriminator network D in a normal GAN outputs an estimated probability that the input image is
drawn from the data generating distribution. Traditionally this is implemented with a feed-forward network
ending in a single sigmoid unit, but it can also be implemented with a softmax output layer with one unit
for each of the classes [REAL, FAKE]. Once this modification is made, it's simple to
see that D could have N+1 output units corresponding to [CLASS-1, CLASS-2, \dots CLASS-N, FAKE].
In this case, D can also act as C. We call this network D/C.

Training an SGAN is similar to training a GAN.
We simply use higher granularity labels for the half of the minibatch that has been drawn from the data generating distribution.
D/C is trained to minimize the negative log likelihood with respect to the given labels and G is trained to maximize it, as shown in Algorithm \ref{alg:example}.
We did not use the modified objective trick described in Section 3 of Goodfellow et al. \yrcite{GANS}.

Note: in concurrent work, \cite{IMPROVEDTECHNIQUES} propose the same method for augmenting
the discriminator and perform a much more thorough experimental evaluation of the technique.

\begin{algorithm}[tb]
   \caption{SGAN Training Algorithm}
   \label{alg:example}
\begin{algorithmic}
   \STATE {\bfseries Input:} $I$: number of total iterations
   \FOR{$i=1$ {\bfseries to} $I$}
   \STATE Draw $m$ noise samples $\{z^{(1)}, \dots, z^{(m)}\}$ from noise prior $p_g(z)$.
   \STATE Draw $m$ examples $\{ (x^{(1)}, y^{(1)}) , \dots, (x^{(m)}, y^{(m)}) \}$ from data generating distribution $p_d(x)$.
   \STATE Perform gradient descent on the parameters of D w.r.t. the NLL of D/C's outputs on the combined minibatch of size $2m$.
   \STATE Draw $m$ noise samples $\{z^{(1)}, \dots, z^{(m)}\}$ from noise prior $p_g(z)$.
   \STATE Perform gradient descent on the parameters of G w.r.t. the NLL of D/C's outputs on the minibatch of size $m$.
   \ENDFOR
\end{algorithmic}
\end{algorithm}

\section{Results}

The experiments in this paper were conducted with \href{https://github.com/DoctorTeeth/supergan}{https://github.com/DoctorTeeth/supergan}, which borrows heavily from \href{https://github.com/carpedm20/DCGAN-tensorflow}{https://github.com/carpedm20/DCGAN-tensorflow} and which contains more details about the experimental setup.

\subsection{Generative Results}

\begin{figure}[ht]
\vskip 0.2in
%\begin{center}
\begin{minipage}[b]{0.5\linewidth}
%\centering
\centerline{\includegraphics[width=\columnwidth]{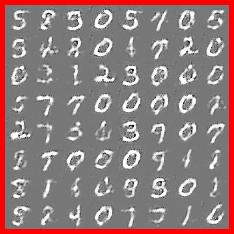}}
\end{minipage}%%
\begin{minipage}[b]{0.5\linewidth}
%\centering
\centerline{\includegraphics[width=\columnwidth]{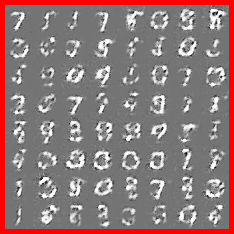}}
\end{minipage} 
\caption{Output samples from SGAN and GAN after 2 MNIST epochs.
SGAN is on the left and GAN is on the right.}
\label{icml-historical}
%\end{center}
\vskip -0.2in
\end{figure}

We ran experiments on the MNIST dataset \cite{MNIST} to determine whether an SGAN would result in better generative samples
than a regular GAN.
Using an architecture similar to that in Radford et al. \yrcite{DCGAN}, we
trained an SGAN both using the actual MNIST labels and with only the labels REAL and FAKE.
Note that the second configuration is semantically identical to a normal GAN.
Figure \ref{icml-historical} contains examples of generative outputs from both GAN and SGAN.
The SGAN outputs are significantly more clear than the GAN outputs.
This seemed to hold true across different initializations and network architectures,
but it is hard to do a systematic evaluation of sample quality for varying hyperparameters.

\subsection{Classifier Results} 
 
We also conducted experiments on MNIST to see whether the classifier component of the SGAN would
perform better than an isolated classifier on restricted training sets.
To train the baseline, we train SGAN without ever updating G.
SGAN outperforms the baseline in proportion to how much we shrink the training set,
suggesting that forcing D and C to share weights improves data-efficiency.
Table \ref{sample-table} includes detailed performance numbers.
To compute accuracy, we took the maximum of the outputs not corresponding to the FAKE label.
For each model, we did a random search on the learning rate and reported the best result.

\begin{table}[t]
\caption{Classifier Accuracy}
\label{sample-table}
\vskip 0.15in
\begin{center}
\begin{small}
\begin{sc}
\begin{tabular}{lcccr}
\hline
\abovespace\belowspace
Examples & CNN & SGAN\\
\hline
\abovespace
1000  &  0.965 & 0.964\\
100   &  0.895 & 0.928\\
50    &  0.859 & 0.883\\
25    &  0.750 & 0.802\\
\belowspace
\end{tabular}
\end{sc}
\end{small}
\end{center}
\vskip -0.1in
\end{table}

\section{Conclusion and Future Work} 
We are excited to explore the following related ideas:

\begin{itemize}
\item Share some (but not all) of the weights between D and C, as in the dual autoencoder \cite{TPUL}. This could allow some weights to be specialized to discrimination and some to classification.
\item Make GAN generate examples with class labels \cite{CONDITIONAL}. Then ask D/C to assign one of $2N$ labels [REAL-ZERO, FAKE-ZERO, \dots ,REAL-NINE, FAKE-NINE].
\item Introduce a ladder network \cite{LADDER} L in place of D/C, then use samples from G as unlabeled data to train L with.
\end{itemize}

\section*{Acknowledgements} 
 
We thank the authors of Tensorflow \cite{TENSORFLOW}.

\bibliography{example_paper}
\bibliographystyle{icml2016}

\end{document}